%% file: main.tex
\documentclass[11pt,a4paper]{article}
\usepackage[utf8]{inputenc}
\usepackage{emnlp2018}
\usepackage{times}
\usepackage{latexsym}
\usepackage{tabularx}

\aclfinalcopy 

\input{defs}

\title{The Importance of Generation Order in Language Modeling}
\author{
    Nicolas Ford\thanks{~Work done as a member of the Google AI Residency program (g.co/airesidency)} \hspace*{0.75cm} Daniel Duckworth \hspace*{0.75cm} Mohammad Norouzi \hspace*{0.75cm} George E.~Dahl\\
    Google Brain\\
    {\tt \{nicf,duckworthd,mnorouzi,gdahl\}@google.com}
}
\date{}

\begin{document}

\maketitle
\begin{abstract}

Neural language models are a critical component of state-of-the-art systems
for machine translation, summarization, audio transcription, and other tasks.
These language models are almost universally autoregressive in
nature, generating sentences one token at a time from left to right. This paper
studies the influence of token generation order on model quality via a novel
\emph{two-pass language model} that produces partially-filled sentence
``templates'' and then fills in missing tokens. We compare various strategies for
structuring these two passes and observe a surprisingly large variation in
model quality. We find the most effective strategy generates {\em function}
words in the first pass followed by {\em content} words in the
second.  We believe these experimental results justify a more extensive
investigation of generation order for neural language models.
\end{abstract}

\section{Introduction}

Neural networks have been extremely successful statistical models of text in language modeling and machine translation. Despite differences in model architectures, state of the art neural nets generate sequences from left to right \citep{attention,jozefowicz2016exploring,wu2016google}. Although in some sense humans produce and consume language from left to right as well, there are many other intuitively appealing ways to generate text. For instance, language is slow enough on a neurological time scale for multiple passes of generation that incorporate feedback to occur. Linguistic intuition might suggest that we should first generate some abstract representation of what we want to say and then serialize it, a process that seems more universally appropriate given the existence of languages with freer word order such as Czech and Polish.


There has been interest in moving beyond the left-to-right generation order by developing alternative multi-stage strategies such as syntax-aware neural language models \citep{bowman2016fast} and latent variable models of text \citep{wood2011sequence}.
Before embarking on a long-term research program to find better generation strategies that improve modern neural networks, one needs evidence that the generation strategy can make a large difference.
This paper presents one way of isolating the generation strategy from the general neural network design problem.
Our key technical contribution involves developing a flexible and \emph{tractable} architecture that incorporates different generation orders, while enabling \emph{exact} computation of the log-probabilities of a sentence.
Our experiments demonstrate that even when using a few simple two-pass generation orders, the differences between good and bad orderings are substantial.

We consider ways of reordering the tokens within a sequence based on their identities.
The best ordering we tried generates function words first and content words last,
which cuts against the idea of committing to the general topic of a sentence first and only then deciding exactly how to phrase it.
We offer some possible explanations in Section~\ref{sec:discussion}, and we conclude that
our experimental results justify a more extensive investigation of the generation order for language and translation models.

\section{Two-pass Language Models}

\begin{table*}[t]
    \centering
    \scriptsize
    \begin{tabularx}{\linewidth}{X|X|X|X|X|X}
{\bf sentence} & {\bf common first} & {\bf rare first} & {\bf function first} & {\bf content first} & {\bf odd first} \\\hline
" all you need to do if you want the nation 's press camped on your doorstep is to say you once had a [UNK] in 1947 , " he noted memorably in his diary . [EOS] & " all you \_\_ to \_\_ if you \_\_ the \_\_ 's \_\_ \_\_ on \_\_ \_\_ is to \_\_ you \_\_ had a [UNK] in \_\_ , " he \_\_ \_\_ in his \_\_ . [EOS] & \_\_ \_\_ \_\_ need \_\_ do \_\_ \_\_ want \_\_ nation \_\_ press camped \_\_ your doorstep \_\_ \_\_ say \_\_ once \_\_ \_\_ \_\_ \_\_ 1947 \_\_ \_\_ \_\_ noted memorably \_\_ \_\_ diary \_\_ [EOS] & " all you \_\_ to \_\_ if you \_\_ the \_\_ 's \_\_ \_\_ on your \_\_ is to \_\_ you \_\_ \_\_ a \_\_ in \_\_ , " he \_\_ \_\_ in his \_\_ . [EOS] & \_\_ \_\_ \_\_ need \_\_ do \_\_ \_\_ want \_\_ nation \_\_ press camped \_\_ \_\_ doorstep \_\_ \_\_ say \_\_ once had \_\_ [UNK] \_\_ 1947 \_\_ \_\_ \_\_ noted memorably \_\_ \_\_ diary \_\_ [EOS] & " all you need \_\_ \_\_ \_\_ you \_\_ the nation 's press camped on your doorstep \_\_ \_\_ say you once had \_\_ \_\_ \_\_ \_\_ \_\_ " \_\_ noted \_\_ \_\_ his \_\_ . [EOS]
\\\hline
the team announced thursday that the 6-foot-1 , [UNK] starter will remain in detroit through the 2013 season . [EOS] & the \_\_ \_\_ \_\_ that the \_\_ , [UNK] \_\_ will \_\_ in \_\_ \_\_ the \_\_ \_\_ . [EOS] & \_\_ team announced thursday \_\_ \_\_ 6-foot-1 \_\_ \_\_ starter \_\_ remain \_\_ detroit through \_\_ 2013 season \_\_ [EOS] & the \_\_ \_\_ \_\_ that the \_\_ , \_\_ \_\_ will \_\_ in \_\_ through the \_\_ \_\_ . [EOS] & \_\_ team announced thursday \_\_ \_\_ 6-foot-1 \_\_ [UNK] starter \_\_ remain \_\_ detroit \_\_ \_\_ 2013 season \_\_ [EOS] & the team announced \_\_ \_\_ the 6-foot-1 \_\_ \_\_ \_\_ will remain \_\_ \_\_ through the 2013 \_\_ . [EOS]
\\\hline
scotland 's next game is a friendly against the czech republic at hampden on 3 march . [EOS] & \_\_ 's \_\_ \_\_ is a \_\_ \_\_ the \_\_ \_\_ at \_\_ on \_\_ \_\_ . [EOS] & scotland \_\_ next game \_\_ \_\_ friendly against \_\_ czech republic \_\_ hampden \_\_ 3 march \_\_ [EOS] & \_\_ 's \_\_ \_\_ is a \_\_ against the \_\_ \_\_ at \_\_ on \_\_ \_\_ . [EOS] & scotland \_\_ next game \_\_ \_\_ friendly \_\_ \_\_ czech republic \_\_ hampden \_\_ 3 march \_\_ [EOS] & \_\_ 's next game \_\_ \_\_ \_\_ \_\_ the czech republic at hampden on 3 march . [EOS]
\\\hline
of course , millions of additional homeowners did make a big mistake : they took advantage of " liar loans " and other [UNK] deals to buy homes they couldn 't afford . [EOS] & of \_\_ , \_\_ of \_\_ \_\_ \_\_ \_\_ a \_\_ \_\_ : they \_\_ \_\_ of " \_\_ \_\_ " and \_\_ [UNK] \_\_ to \_\_ \_\_ they \_\_ 't \_\_ . [EOS] & \_\_ course \_\_ millions \_\_ additional homeowners did make \_\_ big mistake \_\_ \_\_ took advantage \_\_ \_\_ liar loans \_\_ \_\_ other \_\_ deals \_\_ buy homes \_\_ couldn \_\_ afford \_\_ [EOS] & of \_\_ , \_\_ of \_\_ \_\_ \_\_ \_\_ a \_\_ \_\_ : they \_\_ \_\_ of " \_\_ \_\_ " and \_\_ \_\_ \_\_ to \_\_ \_\_ they \_\_ \_\_ \_\_ . [EOS] & \_\_ course \_\_ millions \_\_ additional homeowners did make \_\_ big mistake \_\_ \_\_ took advantage \_\_ \_\_ liar loans \_\_ \_\_ other [UNK] deals \_\_ buy homes \_\_ couldn 't afford \_\_ [EOS] & of \_\_ \_\_ \_\_ of additional \_\_ \_\_ \_\_ \_\_ big \_\_ \_\_ they \_\_ advantage of " liar \_\_ " and other \_\_ deals \_\_ buy homes they couldn \_\_ afford . [EOS]
\end{tabularx}
    \caption{Some example sentences from the dataset and their corresponding templates. The placeholder token is indicated by ``\_\_''.}
    \label{tab:examples}
\end{table*}

We develop a family of {\em two-pass language models} that depend on a partitioning of the vocabulary into a set of \emph{first-pass} and \emph{second-pass} tokens to generate sentences. We perform a preprocessing step on each sequence $\by$, creating two new sequences $\by^{(1)}$ and $\by^{(2)}$. The sequence $\by^{(1)}$, which we call the \emph{template}, has the same length as $\by$, and consists of the first-pass tokens from $\by$ together with a special \emph{placeholder} token wherever $\by$ had a second-pass token. The sequence $\by^{(2)}$ has length equal to the number of these placeholders, and consists of the second-pass tokens from $\by$ in order.

We use a neural language model $p_1$ to generate $\by^{(1)}$, and then a conditional translation model $p_2$ to generate $\by^{(2)}$ given $\by^{(1)}$. Note that, since the division of the vocabulary into first- and second-pass tokens is decided in advance, there is a one-to-one correspondence between sequences $\by$ and pairs $(\by^{(1)},\by^{(2)})$. The total probability of $\by$ is then
\begin{equation}
    p(\by) ~=~ p_{1}(\by^{(1)}) \: p_{2}(\by^{(2)} \mid \by^{(1)})~.
\end{equation}
Two-pass language models present a unique opportunity to study the importance of generation order because, since the template is a deterministic function of $\by$, the probability of $\by$ can be computed exactly. This is in contrast to a language model using a {\em latent} generation order, which requires a prohibitive marginalization over permutations to compute the exact probabilities. Given the tractable nature of the model, exact learning based on log-likelihood is possible, and we can compare different vocabulary partitioning strategies both against each other and against a single-pass language model.

Our implementation consists of two copies of the Transformer model from \citet{attention}. The first copy just generates the template, so it has no encoder. The second copy is a sequence-to-sequence model that translates the template into the complete sentence. There are three places in this model where word embeddings appear --- the first-phase decoder, the second-phase encoder, and the second-phase decoder --- and all three sets of parameters are shared. The output layer also shares the embedding parameters.\footnote{This behavior is enabled in the publicly available implementation of Transformer using the hyperparameter called \texttt{shared\_embedding\_and\_softmax\_weights}.}

For the second pass, we include the entire target sentence, not just the second-pass tokens, on the output side. In this way, when generating a token, the decoder is allowed to examine all tokens to the left of its position. However, only the second-pass tokens count toward the loss, since in the other positions the correct token is already known. Our loss function is then the sum of all of these numbers (from both copies) divided by the length of the original sentence, which is the log-perplexity that our model assigns to the sentence.



We tried five different ways of splitting the vocabulary:

\textbf{Common First and Rare First}: The vocabulary was sorted by frequency and then a cutoff was chosen, splitting the vocabulary into ``common'' and ``rare'' tokens. The location of the cutoff\footnote{In our experiments on LM1B, this is at index 78.} was chosen so that the number of common tokens and the number of rare tokens in the average sentence were approximately the same. In ``common first'' we place the common tokens in the first pass, and in ``rare first'' we start with the rare tokens.

\textbf{Function First and Content First}: We parsed about 1\% of LM1B's training set using Parsey McParseface \citep{parsey} and assigned each token in the vocabulary to the grammatical role it was assigned most frequently by the parser. We used this data to divide the vocabulary into ``function'' words and ``content'' words; punctuation, adpositions, conjunctions, determiners, pronouns, particles, modal verbs, ``wh-adverbs'' (Penn part-of-speech tag \texttt{WRB}), and conjugations of ``be'' were chosen to be function words. In ``function first'' we place the function words in the first phase and in ``content first'' we start with the content words.
    
\textbf{Odd First}: As a control, we also used a linguistically meaningless split where tokens at an odd index in the frequency-sorted vocabulary list were assigned to the first pass and tokens with an even index were assigned to the second pass.

A few sentences from the dataset are shown in Table~\ref{tab:examples} together with their templates. Note that the common and function tokens are very similar; the main differences are the ``unknown'' token, conjugations of ``have,'' and some prepositions.

\section{Experimental Results and Discussion}
\label{sec:discussion}

\begin{table*}[t]
    \centering
    \begin{tabular}{c|c|c|c}
        Model & Train & Validation & Test \\\hline
        odd first & 39.925 & 45.377 & 45.196 \\
        rare first & 38.283 & 43.293 & 43.077 \\
        content first & 38.321 & 42.564 & 42.394 \\
        common first & 36.525 & 41.018 & 40.895 \\
        function first & 36.126 & 40.246 & 40.085 \\
        \hline
        baseline & 38.668 & 41.888 & 41.721 \\
        enhanced baseline & 35.945 & 39.845 & 39.726
    \end{tabular}
    \caption{The perplexities achieved by the best version of each of our models.}
    \label{tab:results}
\end{table*}

We ran experiments with several different ways of splitting the vocabulary into first-pass and second-pass tokens. We trained all of these models on the One Billion Word Language Modeling benchmark (LM1B) dataset \citep{LM1B}. One sixth of the training data was used as a validation set. We used a vocabulary of size 65,536 consisting of whole words (rather than word pieces) converted to lower-case.

We compared the two-pass generation strategies to a baseline version of Transformer without an encoder, which was trained to unconditionally predict the target sentences in the ordinary way. Because the two-pass models contain slightly more trainable parameters than this baseline, we also compare to an ``enhanced baseline'' in which the size of Transformer's hidden space was increased to make the number of parameters match the two-pass models.

Both the two-pass models and the baselines used the hyperparameters referred to as \texttt{base} in the publicly available implementation of Transformer,\footnote{\href{https://github.com/tensorflow/tensor2tensor}{github.com/tensorflow/tensor2tensor}} which has a hidden size of 512, a filter size of 2048, and 8 attention heads, except that the enhanced baseline used a hidden size of 704. We used a batch size of 4096. All models were trained using ADAM \citep{kingma2014adam}, with $\beta_1=0.85$, $\beta_2=0.997$, and $\epsilon=10^{-6}$. The learning rate was tuned by hand separately for each experiment and the experiments that produced the best results on the validation set are reported. Dropout was disabled after some initial experimentation found it to be detrimental to the final validation loss.

Table~\ref{tab:results} shows the results for all the two-pass generation strategies we tried as well as the baselines, sorted from worst to best on the validation set. Strikingly, the linguistically meaningless odd first generation strategy that splits words arbitrarily between the two phases is far worse than the baseline, showing that the two-pass setup on its own provides no inherent advantage over a single phase. The common first and closely related function first strategies perform the best of all the two-pass strategies, whereas the rare first and closely related content first strategies are much worse. Since the control, rare first, and content first orderings are all worse than the baseline, the gains seen by the other two orderings cannot be explained by the increase in the number of trainable parameters alone.

The enhanced version of the baseline achieved slightly better perplexity than the best of the two-pass models we trained. Given that state-of-the-art results with Transformer require models larger than the ones we trained, we should expect growing the embedding and hidden size to produce large benefits. However, the two-pass model we proposed in this work is primarily a tool to understand the importance of sequence generation order and was not designed to be parameter efficient. Thus, as these results indicate, increasing the embedding size in Transformer is a more effective use of trainable parameters than having extra copies of the other model parameters for the second pass (recall that the embeddings are shared across both passes).

One potential explanation for why the function first split performed the best is that, in order to generate a sentence, it is easier to first decide something about its syntactic structure. If this is the primary explanation for the observed results, then common first's success can be attributed to how many function words are also common. However, an alternative explanation might simply be that it is preferable to delay committing to a rare token for as long as possible as all subsequent decisions will then be conditioning on a low-probability event. This is particularly problematic in language modeling where datasets are too small to cover the space of all utterances. We lack sufficient evidence to decide between these hypotheses and believe further investigation is necessary.

Ultimately, our results show that content-dependent generation orders can have a surprisingly large effect on model quality. Moreover, the gaps between different generation strategies can be quite large.

\section{Related Work}
\label{sec:relatedwork}


For tasks conditioning on sequences and sets, it is well known that order significantly affects model quality in applications such as machine translation \citep{seq2seq}, program synthesis \citep{vinyals}, and text classification \citep{yogatama2016learning}. Experimentally, \citet{khandelwal2018sharp} show that recurrent neural networks have a memory that degrades with time. Techniques such as attention \citep{bahdanau2014neural} can be seen as augmenting that memory.

Text generation via neural networks, as in language models and machine translation, proceeds almost universally left-to-right \citep{jozefowicz2016exploring, seq2seq}. This is in stark contrast to phrase-based machine translation systems \citep{charniak2003syntax} which traditionally split token translation and ``editing'' (typically via reordering) into separate stages. This line of work is carried forward in Post-Editing Models \citep{junczys2016log}, Deliberation Networks \citep{xia2017deliberation}, and Review Network \citep{yang2016review} which produce a ``draft'' decoding that is further edited. As any valid sequence may be used in a draft, calculating perplexity in these models is unfortunately intractable, and model quality can only be evaluated via external tasks.

In addition to surface-form intermediate representation, syntax-based representations have a rich history in text modeling. \citet{chelba1998exploiting, yamada2001syntax, graham2010deep, shen2018neural} integrate parse structures, explicitly designed or automatically learned, into the decoding process.

Similar to the second phase of this work's proposed model, \citep{fedus2018maskgan} directly tackles the problem of filling in the blank, akin to the second stage of our proposed model. The Multi-Scale version of PixelRNN in \citep{van2016pixel} was also an inspiration for the two-pass setup we used here.

\section{Conclusion and Future Work}
\label{sec:conclusion}

To investigate the question of generation order in language modeling, we proposed a model that generates a sentence in two passes, first generating tokens from left to right while skipping over some positions and then filling in the positions that it skipped. We found that the decision of which tokens to place in the first pass had a strong effect.



Given the success of our function word first generation procedure, we could imagine taking this idea beyond splitting the vocabulary. One could run a parser on each sentence and use the resulting tree to decide on the generation order. Such a scheme might shed light on which aspect of this split was most helpful. Finally, filling in a template with missing words is a task that might be interesting in its own right. One might want to provide partial information about the target sentence as part of scripting flexible responses for a dialogue agent, question answering system, or other system that mixes a hand-designed grammar with learned responses.


\bibliographystyle{acl_natbib_nourl}
\bibliography{main}

\end{document}

%% file: defs.tex
\usepackage{amsmath}

\renewcommand{\vec}[1]{\boldsymbol{\mathbf{#1}}}
\def\by{\vec{y}}

%% file: main.bbl
\begin{thebibliography}{23}
\expandafter\ifx\csname natexlab\endcsname\relax\def\natexlab#1{#1}\fi

\bibitem[{Andor et~al.(2016)Andor, Alberti, Weiss, Severyn, Presta, Ganchev,
  Petrov, and Collins}]{parsey}
Daniel Andor, Chris Alberti, David Weiss, Aliaksei Severyn, Alessandro Presta,
  Kuzman Ganchev, Slav Petrov, and Michael Collins. 2016.
\newblock Globally normalized transition-based neural networks.
\newblock \emph{CoRR}, abs/1603.06042.

\bibitem[{Bahdanau et~al.(2014)Bahdanau, Cho, and Bengio}]{bahdanau2014neural}
Dzmitry Bahdanau, Kyunghyun Cho, and Yoshua Bengio. 2014.
\newblock Neural machine translation by jointly learning to align and
  translate.
\newblock \emph{arXiv preprint arXiv:1409.0473}.

\bibitem[{Bowman et~al.(2016)Bowman, Gauthier, Rastogi, Gupta, Manning, and
  Potts}]{bowman2016fast}
Samuel~R Bowman, Jon Gauthier, Abhinav Rastogi, Raghav Gupta, Christopher~D
  Manning, and Christopher Potts. 2016.
\newblock A fast unified model for parsing and sentence understanding.
\newblock In \emph{Proceedings of the 54th Annual Meeting of the Association
  for Computational Linguistics (Volume 1: Long Papers)}, volume~1, pages
  1466--1477.

\bibitem[{Charniak et~al.(2003)Charniak, Knight, and
  Yamada}]{charniak2003syntax}
Eugene Charniak, Kevin Knight, and Kenji Yamada. 2003.
\newblock Syntax-based language models for statistical machine translation.
\newblock In \emph{Proceedings of MT Summit IX}, pages 40--46.

\bibitem[{Chelba and Jelinek(1998)}]{chelba1998exploiting}
Ciprian Chelba and Frederick Jelinek. 1998.
\newblock Exploiting syntactic structure for language modeling.
\newblock In \emph{Proceedings of the 36th Annual Meeting of the Association
  for Computational Linguistics and 17th International Conference on
  Computational Linguistics-Volume 1}, pages 225--231. Association for
  Computational Linguistics.

\bibitem[{Chelba et~al.(2013)Chelba, Mikolov, Schuster, Ge, Brants, and
  Koehn}]{LM1B}
Ciprian Chelba, Tom{\'{a}}{\v{s}} Mikolov, Mike Schuster, Qi~Ge, Thorsten
  Brants, and Phillipp Koehn. 2013.
\newblock One billion word benchmark for measuring progress in statistical
  language modeling.
\newblock \emph{CoRR}, abs/1312.3005.

\bibitem[{Fedus et~al.(2018)Fedus, Goodfellow, and Dai}]{fedus2018maskgan}
William Fedus, Ian Goodfellow, and Andrew~M. Dai. 2018.
\newblock Mask{GAN}: Better text generation via filling in the \verb|_______|.
\newblock In \emph{International Conference on Learning Representations
  (ICLR)}.

\bibitem[{Graham and Genabith(2010)}]{graham2010deep}
Yvette Graham and Josef Genabith. 2010.
\newblock Deep syntax language models and statistical machine translation.
\newblock In \emph{Proceedings of the 4th Workshop on Syntax and Structure in
  Statistical Translation}, pages 118--126.

\bibitem[{Jozefowicz et~al.(2016)Jozefowicz, Vinyals, Schuster, Shazeer, and
  Wu}]{jozefowicz2016exploring}
Rafal Jozefowicz, Oriol Vinyals, Mike Schuster, Noam Shazeer, and Yonghui Wu.
  2016.
\newblock Exploring the limits of language modeling.
\newblock \emph{arXiv preprint arXiv:1602.02410}.

\bibitem[{Junczys-Dowmunt and Grundkiewicz(2016)}]{junczys2016log}
Marcin Junczys-Dowmunt and Roman Grundkiewicz. 2016.
\newblock Log-linear combinations of monolingual and bilingual neural machine
  translation models for automatic post-editing.
\newblock \emph{arXiv preprint arXiv:1605.04800}.

\bibitem[{Khandelwal et~al.(2018)Khandelwal, He, Qi, and
  Jurafsky}]{khandelwal2018sharp}
Urvashi Khandelwal, He~He, Peng Qi, and Dan Jurafsky. 2018.
\newblock Sharp nearby, fuzzy far away: How neural language models use context.
\newblock \emph{arXiv preprint arXiv:1805.04623}.

\bibitem[{Kingma and Ba(2014)}]{kingma2014adam}
Diederik Kingma and Jimmy Ba. 2014.
\newblock Adam: A method for stochastic optimization.
\newblock In \emph{International Conference on Learning Representations}.

\bibitem[{Shen et~al.(2018)Shen, Lin, Huang, and Courville}]{shen2018neural}
Yikang Shen, Zhouhan Lin, Chin{-}wei Huang, and Aaron Courville. 2018.
\newblock Neural language modeling by jointly learning syntax and lexicon.
\newblock In \emph{International Conference on Learning Representations}.

\bibitem[{Sutskever et~al.(2014)Sutskever, Vinyals, and Le}]{seq2seq}
Ilya Sutskever, Oriol Vinyals, and Quoc~V Le. 2014.
\newblock Sequence to sequence learning with neural networks.
\newblock In \emph{Advances in Neural Information Processing Systems (NIPS)},
  pages 3104--3112.

\bibitem[{Van~Oord et~al.(2016)Van~Oord, Kalchbrenner, and
  Kavukcuoglu}]{van2016pixel}
Aaron Van~Oord, Nal Kalchbrenner, and Koray Kavukcuoglu. 2016.
\newblock Pixel recurrent neural networks.
\newblock In \emph{International Conference on Machine Learning}, pages
  1747--1756.

\bibitem[{Vaswani et~al.(2017)Vaswani, Shazeer, Parmar, Uszkoreit, Jones,
  Gomez, Kaiser, and Polosukhin}]{attention}
Ashish Vaswani, Noam Shazeer, Niki Parmar, Jakob Uszkoreit, Llion Jones,
  Aidan~N Gomez, {\L}ukasz Kaiser, and Illia Polosukhin. 2017.
\newblock Attention is all you need.
\newblock In \emph{Advances in Neural Information Processing Systems}, pages
  6000--6010.

\bibitem[{Vinyals et~al.(2016)Vinyals, Bengio, and Kudlur}]{vinyals}
Oriol Vinyals, Samy Bengio, and Manjunath Kudlur. 2016.
\newblock Order matters: Sequence to sequence for sets.
\newblock In \emph{International Conference on Learning Representations
  (ICLR)}.

\bibitem[{Wood et~al.(2011)Wood, Gasthaus, Archambeau, James, and
  Teh}]{wood2011sequence}
Frank Wood, Jan Gasthaus, C{\'e}dric Archambeau, Lancelot James, and Yee~Whye
  Teh. 2011.
\newblock The sequence memoizer.
\newblock \emph{Communications of the ACM}, 54(2):91--98.

\bibitem[{Wu et~al.(2016)Wu, Schuster, Chen, Le, Norouzi, Macherey, Krikun,
  Cao, Gao, Macherey et~al.}]{wu2016google}
Yonghui Wu, Mike Schuster, Zhifeng Chen, Quoc~V Le, Mohammad Norouzi, Wolfgang
  Macherey, Maxim Krikun, Yuan Cao, Qin Gao, Klaus Macherey, et~al. 2016.
\newblock Google's neural machine translation system: Bridging the gap between
  human and machine translation.
\newblock \emph{arXiv preprint arXiv:1609.08144}.

\bibitem[{Xia et~al.(2017)Xia, Tian, Wu, Lin, Qin, Yu, and
  Liu}]{xia2017deliberation}
Yingce Xia, Fei Tian, Lijun Wu, Jianxin Lin, Tao Qin, Nenghai Yu, and Tie-Yan
  Liu. 2017.
\newblock Deliberation networks: Sequence generation beyond one-pass decoding.
\newblock In \emph{Advances in Neural Information Processing Systems}, pages
  1782--1792.

\bibitem[{Yamada and Knight(2001)}]{yamada2001syntax}
Kenji Yamada and Kevin Knight. 2001.
\newblock A syntax-based statistical translation model.
\newblock In \emph{Proceedings of the 39th Annual Meeting on Association for
  Computational Linguistics}, pages 523--530. Association for Computational
  Linguistics.

\bibitem[{Yang et~al.(2016)Yang, Yuan, Wu, Cohen, and
  Salakhutdinov}]{yang2016review}
Zhilin Yang, Ye~Yuan, Yuexin Wu, William~W Cohen, and Ruslan~R Salakhutdinov.
  2016.
\newblock Review networks for caption generation.
\newblock In \emph{Advances in Neural Information Processing Systems}, pages
  2361--2369.

\bibitem[{Yogatama et~al.(2016)Yogatama, Blunsom, Dyer, Grefenstette, and
  Ling}]{yogatama2016learning}
Dani Yogatama, Phil Blunsom, Chris Dyer, Edward Grefenstette, and Wang Ling.
  2016.
\newblock Learning to compose words into sentences with reinforcement learning.
\newblock \emph{arXiv preprint arXiv:1611.09100}.

\end{thebibliography}
